\pdfoutput=1

\documentclass[11pt]{article}

\usepackage[final]{acl}

\usepackage{times}
\usepackage{latexsym}

\usepackage[T1]{fontenc}

\usepackage[utf8]{inputenc}

\usepackage{microtype}

\usepackage{inconsolata}

\usepackage{graphicx}
\usepackage{booktabs}

\newcommand\blfootnote[1]{%
  \begingroup
  \renewcommand\thefootnote{}\footnote{#1}%
  \addtocounter{footnote}{-1}%
  \endgroup
}
\title{Evaluation Sheet for Deep Research: \\ A Use Case for Academic Survey Writing}

\author{\normalsize Israel Abebe Azime$^{1,\ast }$ ,  Tadesse Destaw Belay$^{2,\ast}$, 
Atnafu Lambebo Tonja$^{3,\ast}$\\
\footnotesize
 $^1$ Saarland University,   $^2$ Instituto Politécnico Nacional,  $^3$ MBZUAI
\\}

\begin{document}
\maketitle
\blfootnote{$^\ast$ Equal Contribution.}
\begin{abstract}
Large Language Models (LLMs) powered with argentic capabilities are able to do knowledge-intensive tasks without human involvement. A prime example of this tool is Deep research with the capability to browse the web, extract information and generate multi-page reports. In this work, we introduce an evaluation sheet that can be used for assessing the capability of Deep Research tools. In addition, we selected academic survey writing as a use case task and evaluated output reports based on the evaluation sheet we introduced.  Our findings show the need to have carefully crafted evaluation standards. The evaluation done on OpenAI`s Deep Search and Google's Deep Search in generating an academic survey showed the huge gap between search engines and standalone Deep Research tools, the shortcoming in representing the targeted area.
\end{abstract}

\section{Introduction}
Large Language Models (LLMs) present a transformative evolution in artificial intelligence, particularly their capacity for advanced text generation, reasoning, and analytical tasks \cite{naveed2024comprehensiveoverview}. A notable enhancement in LLM functionalities is the incorporation of the vertical AI agents~\cite{bousetouane2025agentic}. Vertical AI agents are specialized intelligent systems tailored to specific industries, combining domain expertise with real-time adaptability to enhance workflows, perform unassisted tasks and decision-making. One of the notable examples of well integrated AI agents is Deep Research. Deep Research  empowers LLMs to perform in-depth examinations of intricate subjects autonomously by accessing web using search engines~\cite{openai}. While the term ``Deep Search'' emphasizes tool delivering quick, concise, and accurate answers through iterative searching and reasoning, ``Deep Research'' leverages reasoning to search, interpret, and analyze information, producing comprehensive long-form reports that explore complex topics in depth~\cite{openai}. In this paper, we focus mainly on Deep Research tools.

\begin{figure}[!t]
    \centering
    \includegraphics[width=\linewidth]{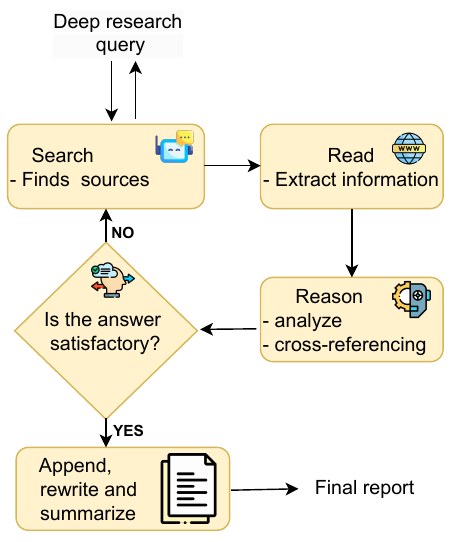}
    \caption{Deep Research workflow}
    \label{fig:drloop}
\end{figure}

Deep Research tools are designed to create comprehensive, long-form reports that dive deep into complex topics~\cite{wu2025agenticreasoning}. Their defining characteristics include unassisted web browsing, compilation of several sources, long waiting time, and results that resemble reports, not chat responses~\cite{openai}.  Deep Research improves traditional search capabilities from keyword-based searching to more exhaustive search incorporating reasoning, inference synthesis, and response generation. This profound research feature transcends basic question-answering; it enables LLMs to navigate the internet, process extensive datasets, synthesize insights, and create structured reports with appropriate citations \cite{xiong2024searchengine}.
Unlike traditional search engines, which primarily provide direct answers, it employs an iterative search process that deconstructs complex inquiries and engages in reasoning before generating responses \cite{wu2025agenticreasoning}. This method operates several search cycles, such as an iterative reading, searching, and reasoning cycle, until the most accurate response is achieved. The entire operation can be segmented into three main distinct phases (search, read and reason), as illustrated in Figure \ref{fig:drloop}.

LLM providers such as Google \footnote{\url{https://blog.google/products/gemini/google-gemini-deep-research/}}, OpenAI \footnote{\url{https://openai.com/index/introducing-deep-research/}}, Perplexity \footnote{\url{https://www.perplexity.ai/ko/hub/blog/introducing-perplexity-deep-research}}, XAI \footnote{\url{https://x.ai/blog/grok-3}} and others are making available their Deep Research agent-based application, a list of ``Deep Research'' tools with their details is shown in Table \ref{tab:deepsearch}. 

One of the main real-world application areas of Deep Research is to use them as helpers for academic research, such as conducting a comprehensive literature review in a specific field of study. Academics can get a draft literature summary in minutes instead of days, and analysts can quickly pull together data from hundreds of webpages. However, these tools still require oversight, and the effectiveness of these Deep Research tools requires rigorous evaluation and study. They might sometimes ``hallucinate'' (produce incorrect information), cite less credible sources or give priority for outdated contents. Even though, Deep Research tools are powerful for scaling up our research capabilities, users must understand their strengths and limitations to choose the right tool. In this work:
\begin{itemize}
    \item We introduce \emph{\textbf{Evaluation Sheet}} as a road-map for evaluating the performance of Deep Research tools. The main categories of this evaluation sheet (\emph{also known as} \textbf{Pillars}) are discussed in Section \ref{pillars}.
    \item As a use case, we selected three recent NLP survey papers focused on African countries and languages: an Ethiopian language survey \cite{tonja-etal-2023-natural}, a Nigerian language survey \cite{inuwadutse2025naijanlpsurvey}, and a Kenyan language survey \cite{amol2024statenlpkenyasurvey} to assess the applicability of the introduced evaluation sheet in order to evaluate the generated Deep Research report. 
    We generated Deep Research reports that resemble these papers from the two selected Deep Research tools (ChatGPT and Gemini) and evaluated their effectiveness, assuming that these survey papers were created using Google-like search engines combined with human involvement. 
\end{itemize}




\begin{table*}[h]
    \centering
    \resizebox{\textwidth}{!}{%
    \begin{tabular}{lllll}
    \toprule
    \textbf{Launch Date} & \textbf{Company} & \textbf{Product} & \textbf{Source} & \textbf{Link} \\
    \midrule

2024-12-11 & Google & Deep Research & Proprietary & \href{https://blog.google/products/gemini/google-gemini-deep-research/}{Google Gemini 2.0} \\
2025-02-02 & OpenAI & Deep Research & Proprietary & \href{https://openai.com/index/introducing-deep-research/}{Introducing Deep Research} \\
2025-02-14 & Perplexity & Deep Research & Proprietary & \href{https://www.perplexity.ai/ko/hub/blog/introducing-perplexity-deep-research}{Introducing Perplexity Deep Research } \\
2025-02-19 & X AI & Grok3 with DeepSearch & Proprietary & \href{https://x.ai/blog/grok-3}{Grok 3 Beta}\\
2025-02-21 & LangChain & Open Deep Research & Open source & \href{https://github.com/langchain-ai/open_deep_research}{Open Deep Research} \\
2025-02-25 & Jina AI & DeepSearch (node-DeepResearch) & Open source & \href{https://search.jina.ai/}{node-DeepResearch | search.jina.ai} \\
2025-03-06 & Manus AI & Manus (beta) & Proprietary & \href{https://manus.im/}{Leave it to Manus}\\
\bottomrule
\end{tabular}
}
\caption{Summaries of Deep Research tools: launch dates in ascending order, company names, products, license types, and source as of March 2025.}
    \label{tab:deepsearch}
\end{table*}

\section{Motivation}
Deep Research
tools allow users to extract, summarize, and gather information on research areas with which they may not be familiar. As the reliability of these tools continues to improve, ensuring their accuracy and dependability is crucial to trusting and using the outputs from these models. LLMs are becoming the new search engines, and if they are not thoroughly tested, research findings may be lost unnoticed, and only selected knowledge will be propagated.

Although Deep Research tools can generate well-structured content, they generate hallucinated references, biased arguments, or incorrect stories. We believe that evaluation sheets are essential to assess AI-generated content to ensure that it meets the required standards, has logical coherence, and has relevant and appropriate sources. This evaluation sheet helps the users to determine whether AI-generated content has accurate data, unbiased output, and diverse perspectives. It also provides ways to verify the source's credibility and the generated text's reliability. The users can effectively leverage Deep Research tools using the evaluation sheet while maintaining the required standards. We also hope that researchers will use this evaluation sheet as a starting point and add more pillars along with questions to create a standard way of testing Deep Research tools.

\section{The Evaluation Sheets - Pillars }
\label{pillars}

LLM evaluation datasets, particularly those focusing on low-resource languages, should emphasize specific characteristics of the generated output. In this work, we propose evaluation sheets that contain different questions in five pillars to evaluate LLMs' Deep Research tool 
that require minimal user interaction. The proposed evaluation sheet can be further adapted and extended to create different benchmark datasets to evaluate different LLM tools for different use cases.  Here, we discuss the six pillars of the proposed evaluation sheet:

\textbf{(1) LLMs \& Deep Research for} [\textit{Surveying NLP Papers and Datasets for Low-Resource African Languages}]\footnote{This section and subsequent questions can be replaced or modified according to the use case scenario (Eg. financial market study, Sport analysis etc).}. Surveying existing NLP papers in research areas such as low-resource languages presents unique challenges. A crucial task is determining whether these tools can effectively identify the most important and impactful research, even when such research papers do not appear in the top search results. The primary issue we aim to address is how the growing popularity of these tools and their increasing role in replacing traditional
searche engines affects the \emph{\textbf{visibility and accessibility of significant research}}.

 To access the usage of LLMs \& Deep Research in survey report writing in low-resource languages, we crafted the following question:   
\begin{itemize}
\item \emph{Does the Deep Research reports effectively identifies and consolidate NLP papers on low-resource [African]\footnote{can be specific region name (Ethiopia, Kenya and Nigeria)} languages?}
\item \emph{Does the selection of datasets for low-resource [African] languages is comprehensive and representative?}
\item \emph{Does the  Deep Research method provide sufficient depth in its analysis of linguistic challenges in [African] NLP?}
\item \emph{Does the  LLM-generated survey highlights the most impactful research in [African] NLP?}
\item \emph{Does the coverage of low-resource [African] languages in the survey align with the actual research landscape?}
\end{itemize}

\textbf{(2) Hallucination}
Hallucination refers to information that appears true to someone without prior knowledge of the subject but cannot be verified by a reliable source~\cite{huang2025survey}. In contrast, errors are categorized as mistakes that are easily noticeable. Hallucination is a huge treat in practical LLM usage, specifically while automating knowledge extraction from contents like research works. This set of guidelines and questions helps us determine the focus we must place on the reliability of the output.
The following questions are crafted to evaluate whether the Deep Research generated report contains hallucination.

\begin{itemize}
\item \emph{Does the  Deep Research generated survey contains minimal factual errors or hallucinations?}
\item \emph{Does the hallucinated content, if present, is easy to identify and correct?} 
\item \emph{Does the  Deep Research tool properly distinguishes between verified academic sources and speculative content?}
\item \emph{Does a lower risk of hallucination improve the reliability of the survey’s insights?}
\end{itemize}

\textbf{(3) Correctness of sources}
Sources can range from reliable, peer-reviewed papers to blogs and social media pages that present personal opinions. While extracting information from both types of sources is optional, web agents should be able to distinguish between reliable and unreliable sources. Below, we pose a set of questions to assess whether the source impacts the reliability of the information and whether certain sources are preferable. This approach ensures that the extracted information is accurate and verified.  

\begin{itemize}
\item \emph{Does the sources suggested in the report are based on verifiable and authoritative sources?}
\item \emph{Does the  Deep Research tool appropriately prioritize papers on credibility and impact?}
\item \emph{Does the mechanism used by Deep Research to extract information from sources adequately account for domain-specific knowledge in [NLP]?}
\end{itemize}

\textbf{(4) Information Validity }
The validity of the references provided can be assessed based on their accessibility, verification through independent sources, and whether they demonstrate why they are superior to other potential alternatives. Below are the questions created to assess the validity of information generated by Deep Research. 

\begin{itemize}
\item \emph{Does the  cited links and references in the survey are valid and accessible?}
\item \emph{Does the  Deep Research tool effectively differentiates between credible and non-credible sources?}
\item \emph{Does the  report content remains valid and relevant when cross-checked with independent sources?}
\item \emph{Does the  Deep Research tool provide sufficient transparency regarding how sources are selected and ranked?}
\item \emph{Does the  Deep Research generated report appropriately handles broken or outdated links in its output?}
\end{itemize}

\textbf{(5) Information Latestness}
Recent information is more valid compared to older information that may have a high search volume but could have been corrected or improved by more recent works. Research papers with higher citation counts and those that appear at the top of search results are not always the latest studies, which can pose a challenge for LLM agents searching the web for information. The following question will help to assess whether the information generated in the report has been extracted from the latest sources. 
\begin{itemize}
\item \emph{Does the report prioritize the most recent sources?}
\item \emph{Does the  Deep Research tool effectively identify the latest trends in NLP for low-resource African languages?}
\item \emph{Does the Deep Research method ensure that outdated references are minimized in the survey?}
\item \emph{Does the system effectively highlight emerging resources that are not widely recognized?}
\item \emph{Does the report output remain relevant given the fast-paced evolution of AI and [NLP] research?}
\end{itemize}

\begin{figure*}[!ht]
    \centering
    \includegraphics[width=7.5cm]{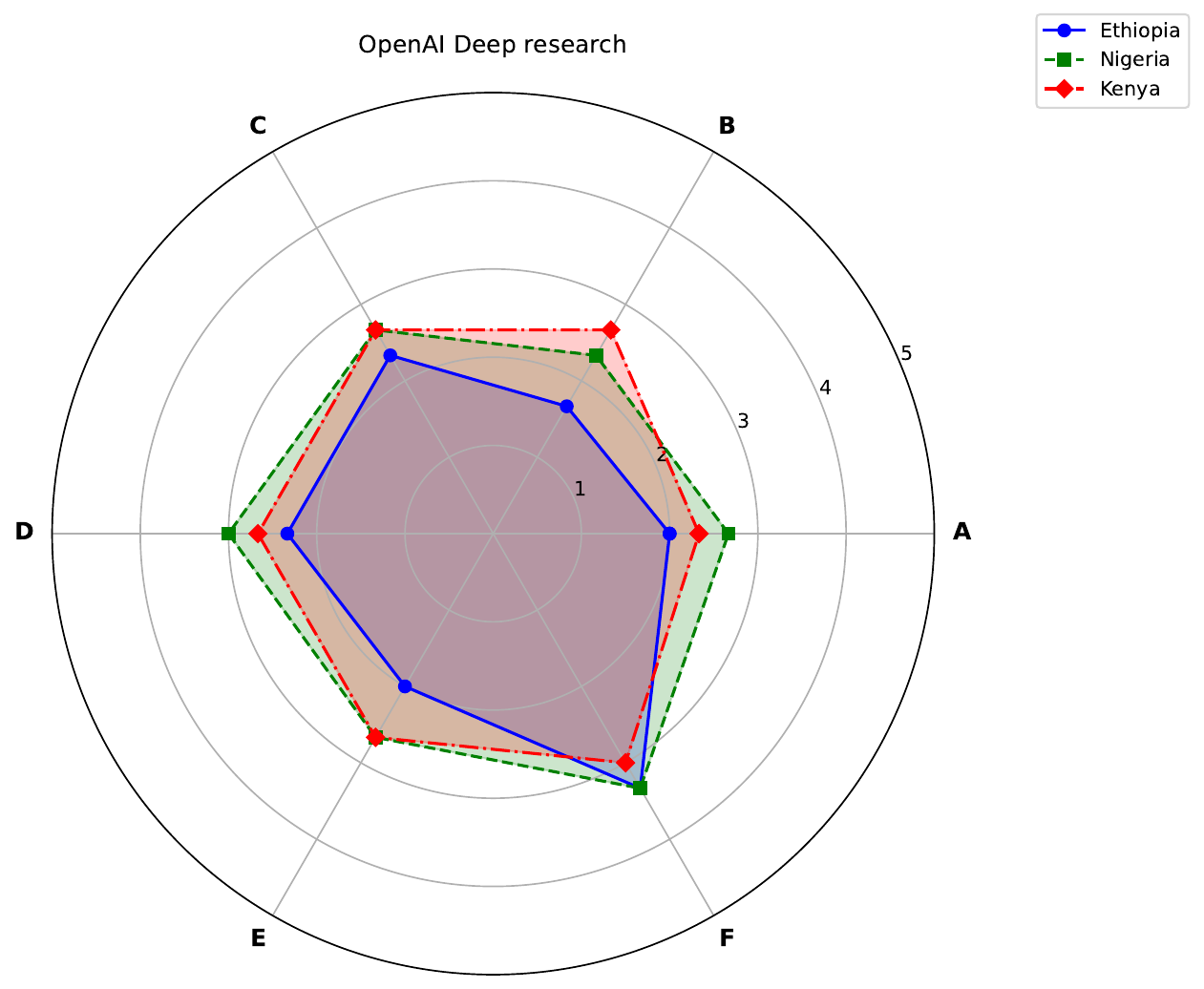}
    \includegraphics[width=7.5cm]{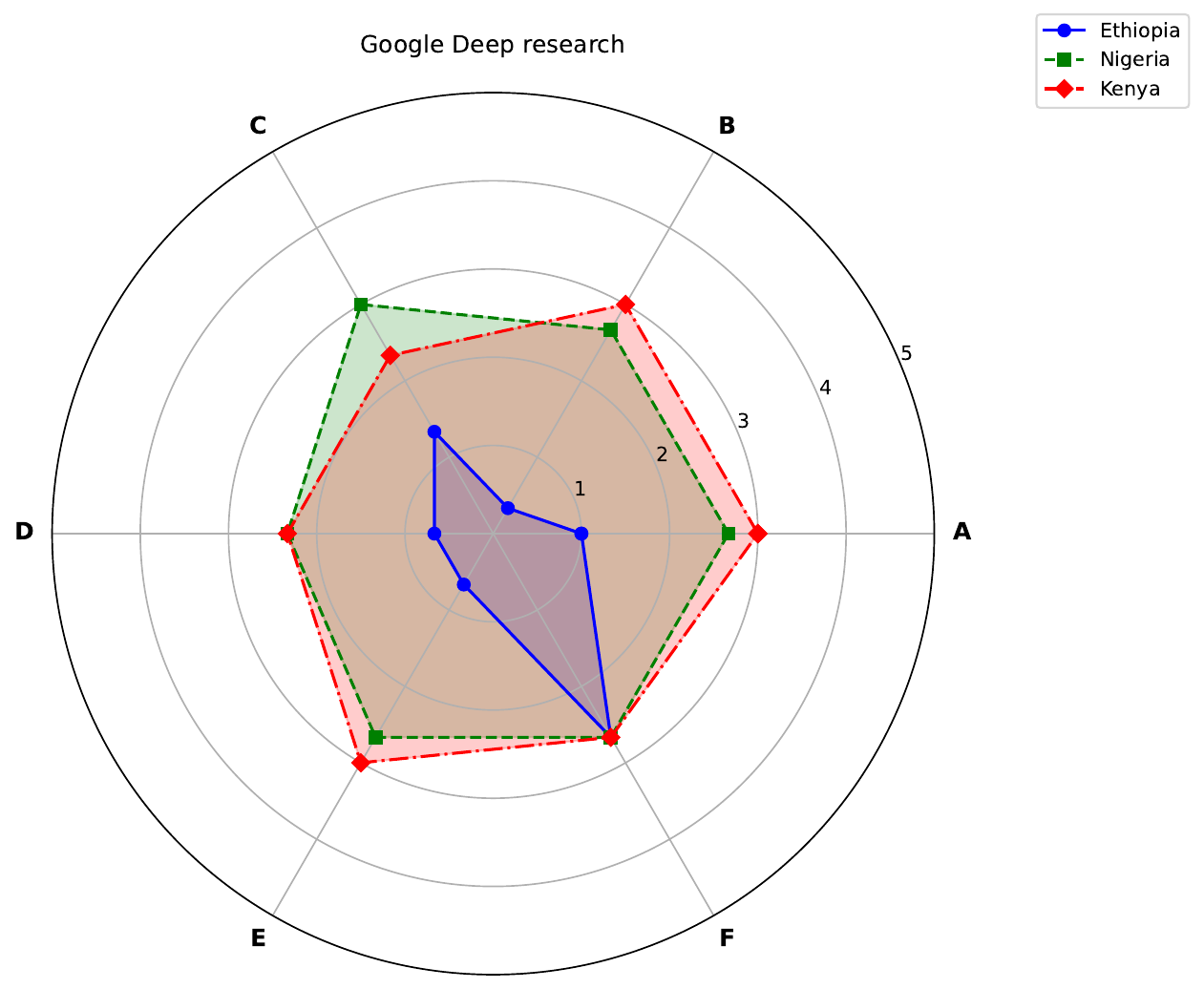}
    \caption{\textbf{A} – LLMs \& Deep Research for Surveying NLP Papers, \textbf{B} – Hallucination,
	\textbf{C} – Correction Sources,
	\textbf{D} – Information/Link Validity,
	\textbf{E} – Information Latestness,
	\textbf{F} – Quantifying Actual Google Search Results vs. LLM Answers,
    }
    \label{fig:results}
\end{figure*}

\textbf{ (6) Quantifying Actual Google Search Results vs. Deep Research Answers}

Finally, we added questions below to explore how the shift from using search engines like Google for information retrieval compares to using automated search agents like Deep Research tools.

\begin{itemize}
\item \emph{Does the report findings align well with actual Google search results on the same topics?}
\item \emph{Does Deep Research generated answers provided by Deep Research are insightful than Google search results?}
\item \emph{Does the  Deep Research tool accurately quantify differences in retrieval efficiency between LLMs and traditional search engines?}
\item \emph{Does the Deep Research tool effectively reduce misinformation compared to open-web search engines?}
\item  \emph{Does the Deep Research approach provide added value beyond standard keyword-based search queries?}
\end{itemize}




\subsection{Rating Procedure}
For the above questions (listed in Section \ref{pillars}), we recommend that users use the Likert scale \cite{joshi2015likert} rating system when answering.  
The rating scale consists of six levels to express agreement or disagreement with a question. These are: \textbf{Strongly Disagree (0)}- indicates complete opposition with no support for the statement. \textbf{Disagree (1)}- reflects mostly disagreement, though some merit is acknowledged. \textbf{Somewhat Disagree (2)}- suggests a leaning toward disagreement while recognizing certain validity. \textbf{Neutral (3)}- signifies neither agreement nor disagreement or an undecided stance. \textbf{Somewhat Agree (4)}- represents general agreement but with some reservations. Finally, \textbf{Strongly Agree (5)}-expresses full endorsement and support without any doubt.


\section{Case study: Ethiopia, Nigeria, Kenya}
\subsection{Methodology}
\paragraph{Creating evaluation sheet}
We selected three regional survey papers that focus on capturing valuable research progress within their respective countries: the Ethiopian language survey \cite{tonja-etal-2023-natural}, the Nigerian language survey \cite{inuwadutse2025naijanlpsurvey}, and the Kenyan language survey \cite{amol2024statenlpkenyasurvey}. We analyzed these papers in detail, extracted the key questions they addressed, and then combined them to formulate prompts (see \ref{prompt})incorporating these questions. To create the evaluation sheet, we carefully identified scenarios the Deep Research tools fail at and must be tested with and created a list of questions under each important evaluation topic. Questions were edited, filtered and removed based on discussion among the authors.

\paragraph{Generating representative outputs}
We evaluated the prompts for validity and selected the one capable of generating detailed reports. Using a selected prompt, we generated three distinct Deep Research outputs by modifying only the country-specific information while utilizing OpenAI Deep Research and Google Deep Research. Three reviewers selected from the authors of this study reviewed the outputs of the tools and rated the generated report based on the rating criteria for each question in the pillars. They used the actual research paper from each of the countries as a reference while answering the questions accordingly.  

\subsection{Comparative analysis}
In this section, we discuss our observations while evaluating reports generated by Google’s Deep Research and OpenAI’s Deep Research tools. Due to the limited number of reports considered in this study and the frequent updates made to these tools, we focus only on the broader conclusions from the results. We recommend scaling this work with a larger number of reports and evaluators to derive more detailed findings. 
\paragraph{LLMs \& Deep Research for Surveying NLP Papers}
Both Google's Deep Research and OpenAI`s Deep Research tools show below-average results in identifying more valuable research works in their reports. The region-specific gap becomes larger for Google's Deep Research.
\paragraph{Hallucination} The inclusion of social media links alongside verified academic peer review catalogs as sources makes Deep Research tools particularly susceptible to hallucinations and erroneous outputs. Additionally, the absence of source information in reports or the citation of incorrect sources complicates the process of identifying and verifying hallucinations. However, based on our analysis, we found that the rate of misinformation and hallucination is not significantly high.

\paragraph{Correctness of Sources}
When examining the detailed process these tools follow while ``researching'', they tend to review a large number of relevant resources. Google’s tool heavily summarizes information and often does not mention many of the sources it picks up during the process. Additionally, both tools tend to include social media links, such as Facebook and Reddit, as information sources.

\paragraph{Information/Link Validity}
We observe that the tools use sources multiple times during their execution. Apart from that, the tools have a problem of identifying the correct source from which the information is obtained and mostly rely on survey papers and summarized contents rather than extracting information from the original source. 

\paragraph{Actual Google Search Results vs. LLM Answers}

Although the system does not produce significant misinformation, its outputs are not fully aligned with Google search results. We find better choices, more recent works, and broader domain coverage when using Google Search.
\subsection{Lesson learned - Takeaway}

\paragraph{The need for evaluation standard} With the rapid introduction of tools that improve or entirely replace search engines, it is crucial to establish evaluation guidelines that foster consistency and common characteristics across benchmarks. The careful design and assessment of these tools are essential, as they shape the knowledge and research considered important, as well as how different approaches and solutions are presented for comparison, ultimately influencing decision-making. If these tools are not designed to provide as much relevant information as possible to users, the real decision-making process—including the selection of problems and solutions—risks being controlled by autonomous agents developed by big tech companies.

\paragraph{Are Deep Research tools reliable for extracting information and generating user-ready reports for low resource research summarization?} The use cases in this study, focused on generating scientific summary reports on underrepresented groups, highlight the challenges of finding, sorting, and presenting hard-to-access research. We found that Deep Research tools are \textbf{not fully reliable}, as their selection of research works lacks transparency, and their summaries—drawn from multiple sources—fail to comprehensively represent the research landscape of the targeted area.

Despite the limitations discussed above, Deep Research tools have a potential in presenting summarized information and making it more accessible.




\section{Conclusion}

LLMs equipped with web search capabilities can delve deeper and spend more time answering questions, making them valuable for knowledge-intensive tasks by comparing multiple sources and improving reasoning. The introduction of Deep Research tools exemplifies this capability, enabling LLMs to search for sources, filter numerous links, and generate detailed reports.

In this work, we developed an Evaluation Sheet to help researchers identify the most critical evaluation criteria for assessing Deep Research tools for different use cases. This evaluation sheet seeks to standardize benchmarking datasets by highlighting key focus areas. To demonstrate its applicability, we conducted a proof-of-concept study on ``Deep Research for Survey Paper Generation'' and used it to evaluate two well-known Deep Research tools.

We hope researchers will adopt this Evaluation Sheet to create benchmarking datasets in their respective domains, ultimately improving the effectiveness of agentic tools that require minimal human interaction. By ensuring these tools generate reliable and informative outputs—comparable to what users would find through independent searches—we aim to improve their practical utility and trustworthiness.

\section*{Limitation}

 Deep Research tools are relatively new, and we selected OpenAI and Google as use cases due to their availability and popularity. Future research will expand the scope by incorporating a broader range of tools, generating a larger number of reports and a larger number of evaluators to better assess their capabilities on a wider scale.

\section*{Acknowledgment}

The authors would like to thank the German Federal Ministry of Education and Research and the German federal states (http://www.nhr-verein.de/en/our-partners) for supporting this work/project as part of the National High-Performance Computing (NHR) joint funding program. 

\bibliography{custom}

\onecolumn
\newpage
\appendix

\section{Prompt}
\label{prompt}

\textbf{Deep Research Template for NLP Survey on a Specific Country}

\textbf{Steps to Conduct This NLP Survey}

\textbf{Step 1: Define Your Research Scope}  
Select the country whose NLP landscape you want to analyze. Identify the languages spoken in the country, including official, regional, indigenous, and endangered languages. Decide on the specific NLP focus, such as general NLP, speech recognition, machine translation, or sentiment analysis.

\textbf{Step 2: Gather Data \& Sources}  
\begin{itemize}
    \item \textbf{Academic Papers}: Search IEEE Xplore, ACL Anthology, Google Scholar, arXiv, and Scopus.
    \item \textbf{Datasets \& Resources}: Explore Hugging Face, Kaggle, LDC, and government data repositories.
    \item \textbf{Pretrained Models}: Check models from Hugging Face, Google AI, and Meta AI.
    \item \textbf{Government \& Industry Reports}: Look for language policy documents and AI research reports.
    \item \textbf{Community \& Open-Source Projects}: Identify ongoing grassroots NLP efforts.
\end{itemize}

\textbf{Step 3: Structure the Paper Using the Template Below}  

Use the structured sections to analyze and organize findings. Answer the guiding questions within each section to provide a comprehensive analysis.

\textbf{Step 4: Conduct Systematic Analysis}  

Review historical NLP progress in the country. Evaluate language challenges and computational constraints affecting NLP adoption. Identify key gaps in linguistic resources, datasets, and models. Highlight ongoing projects and promising research directions.

\textbf{Step 5: Synthesize Findings \& Propose Solutions}  

Summarize research trends, NLP applications, and linguistic barriers. Suggest data collection initiatives, model improvements, and collaborative strategies. Provide policy recommendations for governments, industries, and researchers.

\textbf{Research Template: Structure of the Paper}  
\begin{itemize}
    \item \textbf{Introduction}: Define the research focus, its importance, and the major linguistic and computational challenges in the country.
    \item \textbf{Research Methodology}: Describe the sources used, search strategies, and inclusion/exclusion criteria.
    \item \textbf{Language Landscape}: Analyze linguistic diversity, digital presence, and computational challenges.
    \item \textbf{Available NLP Resources \& Tools}: Review datasets, pretrained models, and language processing tools.
    \item \textbf{NLP Applications \& Downstream Tasks}: Discuss various NLP tasks such as text processing, machine translation, ASR, NER, and conversational AI.
    \item \textbf{Challenges \& Limitations}: Address technical constraints, linguistic barriers, and ethical concerns.
    \item \textbf{Future Directions \& Recommendations}: Propose solutions for data collection, model improvements, policy considerations, and community engagement.
    \item \textbf{Conclusion}: Summarize key findings and provide a call to action.
\end{itemize}

\textbf{Guiding Questions for Each Section}

\textbf{1. Introduction}
\begin{itemize}
    \item What is the focus of this research?
    \item Why is this topic important for [Country Name]?
    \item What are the major linguistic and computational challenges in this country’s NLP landscape?
    \item What are the objectives and scope of this study?
    \item How does the country’s NLP research compare to global trends?
\end{itemize}

\textbf{2. Research Methodology}
\begin{itemize}
    \item What databases and sources were used?
    \item What search strategies were applied?
    \item What criteria were used to include/exclude studies?
    \item How was the information categorized (e.g., by language type, NLP task, dataset availability)?
\end{itemize}

\textbf{3. Language Landscape in [Country Name]}
\begin{itemize}
    \item What are the primary linguistic characteristics of the country’s languages?
    \item Which languages have the most NLP research, and which are neglected?
    \item What challenges arise in processing these languages (e.g., word segmentation, diacritics)?
\end{itemize}

\textbf{4. Available NLP Resources \& Tools}
\begin{itemize}
    \item Are there high-quality datasets available for these languages?
    \item Are the models pre-trained on country-specific linguistic data?
    \item What tools exist for POS tagging, NER, and other NLP tasks?
\end{itemize}

\textbf{5. NLP Applications \& Downstream Tasks}
\begin{itemize}
    \item What NLP tasks have seen the most research focus?
    \item What tools and datasets exist for these tasks?
    \item What are the biggest challenges in implementing NLP solutions?
\end{itemize}

\textbf{6. Challenges \& Limitations}
\begin{itemize}
    \item What are the biggest challenges preventing NLP advancements?
    \item Are there systematic biases in datasets and models?
    \item How does governmental or industry support impact NLP growth?
\end{itemize}

\textbf{7. Future Directions \& Recommendations}
\begin{itemize}
    \item What strategies can bridge the research gap in NLP for [Country Name]?
    \item What government or private sector initiatives can support NLP growth?
    \item How can the NLP community collaborate to improve datasets and models?
\end{itemize}

\textbf{8. Conclusion}  

Summarize key findings and provide a call to action for researchers, policymakers, and industry leaders.

\vspace{1cm}
\vspace{1cm}

\textbf{Practical Example: Applying This Template}  
\begin{itemize}
    \item \textbf{Choose the country}: Kenya.
    \item \textbf{Select the languages}: Swahili (major language), Kikuyu, Luo, Maasai (regional languages).
    \item \textbf{Determine the focus}: Speech recognition \& machine translation.
    \item \textbf{Collect data}: Look for Kenyan NLP research, datasets, and community projects.
    \item \textbf{Analyze findings}: Identify gaps, challenges, and progress in NLP research.
    \item \textbf{Suggest solutions}: Recommend better dataset collection, funding initiatives, and collaborative research.
\end{itemize}


\end{document}